\documentclass[conference]{IEEEtran}
\IEEEoverridecommandlockouts
\usepackage{cite}
\usepackage{svg}
\usepackage{makecell}
\usepackage{amsmath,amssymb,amsfonts}
\usepackage{algorithmic}
\usepackage{graphicx}
\usepackage{textcomp}
\usepackage{xcolor}
\usepackage{hyperref}
\usepackage{arydshln}
\usepackage{booktabs}
\usepackage{multirow}
\newcommand{\model}{KA$^2$LM}
\newcommand{\data}{\textit{Audiopedia}}
\usepackage{pifont}
\newcommand{\cmark}{\ding{51}}%
\newcommand{\xmark}{\ding{55}}%
\def\BibTeX{{\rm B\kern-.05em{\sc i\kern-.025em b}\kern-.08em
    T\kern-.1667em\lower.7ex\hbox{E}\kern-.125emX}}

\begin{document}

\title{Audiopedia: Audio QA with Knowledge
\thanks{\textsuperscript{*} These authors contributed equally to this work.}}

\author{\IEEEauthorblockN{Abhirama Subramanyam Penamakuri\textsuperscript{1*}, Kiran Chhatre\textsuperscript{2*}, Akshat Jain\textsuperscript{1}}
\IEEEauthorblockA{
\textit{\textsuperscript{1}Indian Institute of Technology, Jodhpur  \textsuperscript{2}KTH Royal Institute of Technology, Sweden}\\
penamakuri.1@iitj.ac.in, chhatre@kth.se, jain.73@iitj.ac.in}
}

\maketitle

\begin{abstract}

In this paper, we introduce \data{}, a novel task called Audio Question Answering with Knowledge, which requires both audio comprehension and external knowledge reasoning. Unlike traditional Audio Question Answering (AQA) benchmarks that focus on simple queries answerable from audio alone, \data{} targets knowledge-intensive questions. We define three sub-tasks: (i) Single Audio Question Answering (s-AQA), where questions are answered based on a single audio sample, (ii) Multi-Audio Question Answering (m-AQA), which requires reasoning over multiple audio samples, and (iii) Retrieval-Augmented Audio Question Answering (r-AQA), which involves retrieving relevant audio to answer the question. We benchmark large audio language models (LALMs) on these sub-tasks and observe suboptimal performance. To address this, we propose a generic framework that can be adapted to any LALM, equipping them with knowledge reasoning capabilities. Our framework has two components: (i) Audio Entity Linking (AEL) and (ii) Knowledge-Augmented Audio Large Multimodal Model (\model{}), which together improve performance on knowledge-intensive AQA tasks. To our knowledge, this is the first work to address advanced audio understanding via knowledge-intensive tasks like \data{}. 

\end{abstract}


\section{Introduction}

\begin{figure}[t!]
   \centering
   \scriptsize
  \includegraphics[width=\columnwidth]{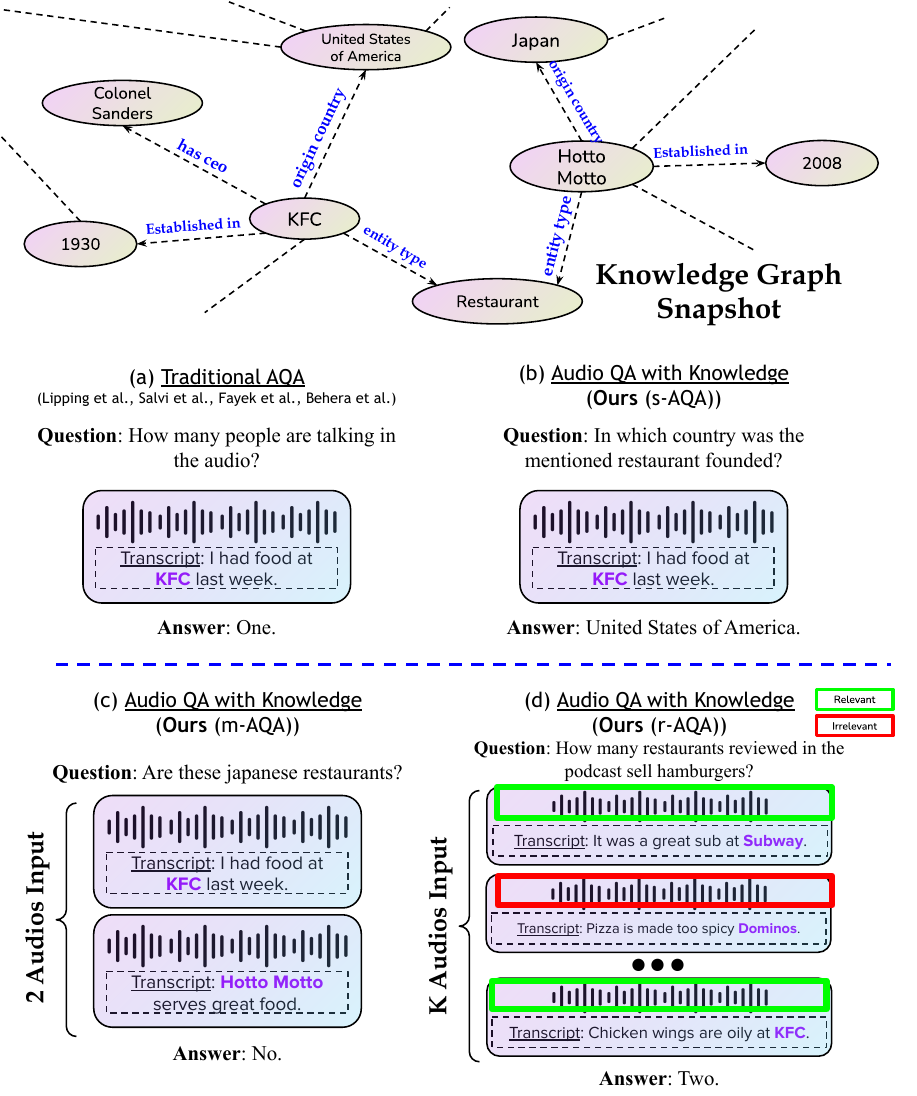}
\caption{\textbf{Visual Abstract of our proposed task and its sub-tasks}: Comparison between traditional AQA (a) and our proposed \data{} sub-tasks: s-AQA (b), m-AQA (c), and r-AQA (d). Unlike traditional AQA, our sub-tasks require reasoning over external knowledge linked to the mentioned named entities (shown at the top). Further details are provided in Section~\ref{sec:dataset}.}
 \label{fig:teaser_fig}
\end{figure}

The research community has shown increasing interest in building chatbots and personal assistants~\cite{gpt3,zhu2023minigpt,dubey2024llama}, driven by the recent success of large language and multimodal models. In this context, Audio Question Answering (AQA)—which involves answering questions based on input audio, enhances these assistants with more advanced interaction capabilities. Recent literature~\cite{lipping2022clotho,salviclear,fayek2020temporal,behera2023towards} has made noticeable progress in AQA, where questions are typically answerable from the audio content. However, these approaches overlook a critical real-world scenario: answering questions about named entities in the audio, which requires reasoning beyond the audio content and drawing on external world knowledge. We refer to this challenge as `\data{}: Audio Question Answering with Knowledge', and such questions as knowledge-intensive questions. An example of traditionally studied AQA vs our proposed \data{} is shown in Fig.~\ref{fig:teaser_fig} (a) and (b), respectively. To accurately answer a knowledge-intensive question such as `In which country was the mentioned restaurant founded?', a system has to understand and reason over the knowledge associated with the mentioned named entity `KFC' from an external knowledge base. While a similar setting has been extensively studied in vision-language research~\cite{shah2019kvqa,schwenk2022okvqa,infoseek,penamakuri2023answer}, it remains unexplored in the audio domain. Moreover, there is no existing dataset for benchmarking methods on knowledge-intensive audio tasks, highlighting the need for such a dataset to advance research in this area. Our work aims to address this gap.

Towards this end, we propose three sub-tasks under the umbrella of \textsc{\data}, to address various requirements of real-world applications inspired from the vision literature~\cite{shah2019kvqa,schwenk2022okvqa,infoseek,imageset_vqa,tanaka2023slidevqa,penamakuri2023answer}. These tasks are as follows: (i) Single audio question answering (s-AQA): where a question has to be answered from the context of a single audio. (ii) Multi audio question answering (m-AQA): where a question has to be answered from the context of multiple audios; this setting is in fact more challenging than the s-AQA setting as in this setting to answer a question, the model has to reason over knowledge associated with named entity mentions across multiple audios. (iii) Retrieval-augmented audio question answering (r-AQA): this is the most challenging task of all. In this setting, a question must be answered using a set of audio clips, where only a few are relevant to the question. To generate a correct answer, the model must retrieve the relevant audios from the pool and reason over the entities across these retrieved clips. For all three sub-tasks, we present a \textit{derived} synthetic evaluation datasets leveraging the knowledge-base of TextKVQA~\cite{singh2019strings}. More details on dataset curation are in Section~\ref{sec:dataset}. 

Further, inspired from~\cite{vistel}, we propose a generic framework that can enhance any LALM with knowledge reasoning capabilities. Our proposed framework has two key components: (i) \underline{A}udio \underline{E}ntity \underline{L}inking (AEL) which identifies and links the named entity mentions in the audio to their associate knowledge in the knowledge base, and (ii) \underline{K}nowledge-\underline{A}ugmented \underline{A}udio \underline{L}arge \underline{M}ultimodal model (\model{}) where we augment the LALM with the additional knowledge obtained through AEL. Owing to its simplicity, our proposed framework can be integrated into any LALM. We conduct extensive experiments using three recent LALMs, specifically Audio-flamingo~\cite{audioflamingo}, GAMA~\cite{ghosh2024gama}, and LTU-AS~\cite{ltu}, demonstrating the superior performance achieved over all three \data{} sub-tasks by incorporating our proposed framework. We make our data publicly available.\footnote{\href{https://github.com/Abhiram4572/Audiopedia}{https://github.com/Abhiram4572/Audiopedia}}


\begin{figure*}[t!]
   \centering
   \scriptsize
  \includegraphics[width=\textwidth]{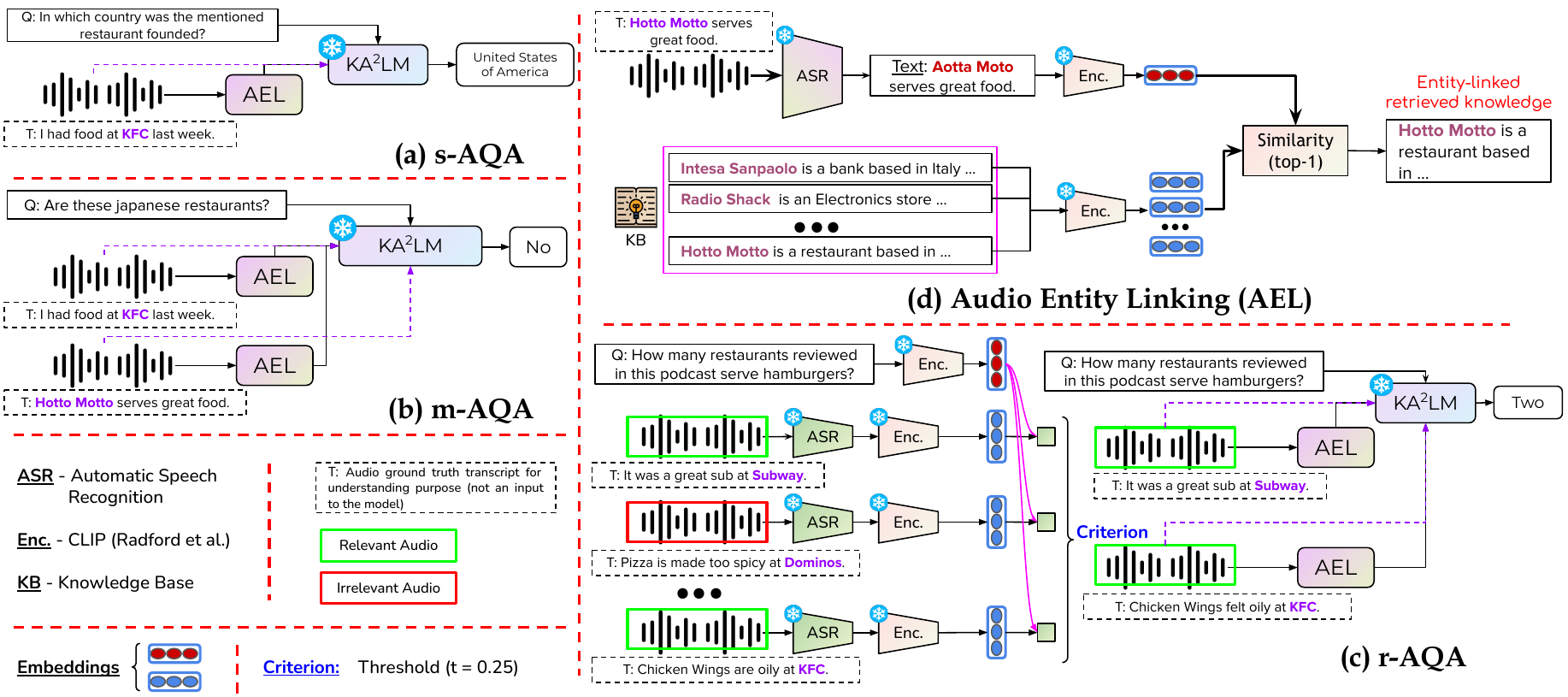}
\caption{{\textbf{An overview of our proposed methodology}: (a) s-AQA: Given an audio, we first perform AEL to obtain the knowledge corresponding to the named entity mentioned in the audio; which is further leveraged by \model{} to augment a LALM to arrive at an accurate answer. (b) m-AQA: Given multiple audios, we perform AEL on every input audio to obtain their associated knowledge; which is fed to \model{} along with concatenated input audios, to generate an accurate answer, (c) r-AQA: We solve this task by retrieving relevant audios from the set of the audios first; once retrieved an accurate answer is obtained in a similar fashion as (b). (d) AEL method: We pose the audio entity linking as a ranking problem. Detailed explanation in Section.~\ref{sec:method}.}}
 \label{fig:main_arch}
\end{figure*}


\begin{table}[!t]
\begin{minipage}[t]{0.30\columnwidth}
\centering
    \caption{s-AQA stats}
    \resizebox{\columnwidth}{!}{
    \begin{tabular}{ll}
    \hline
    Measurement & Value \\
    \midrule
    \#Samples & 14702 \\
    Answer type & Open-ended \\
    \#Unique answers & 1039 \\     
    \hline
    \end{tabular}}
  \label{minitab:s-AQA stats}
\end{minipage}
\begin{minipage}[t]{0.25\columnwidth}
\centering
    \caption{m-AQA stats}
    \resizebox{\columnwidth}{!}{
   \begin{tabular}{ll}
   \hline
   Measurement & Value \\
   \midrule
   \#Samples & 500 \\
   Answer type & Binary \\
   \hline
   \end{tabular}}
  \label{minitab:m-AQA stats}
\end{minipage}
\begin{minipage}[t]{0.4\columnwidth}
\centering
    \caption{r-AQA stats}
    \resizebox{\columnwidth}{!}{
   \begin{tabular}{ll}
   \hline
   Measurement & Value\\
   \midrule
   \#Samples & 115 \\
   Answer type & Counts\\
   {\#Avg. relevant APQ} & 2.2\\
   {\#Avg. irrelevant APQ} & 8.3\\
   \hline
   {\centering \scriptsize{(APQ: Audios per question)}}
   \end{tabular}}
  \label{minitab:r-AQA stats}
\end{minipage}
\end{table}

\section{Related Work}
Question answering is a task to answer the natural language question over the given inputs. Depending on the modality of the input, question answering in the literature can be categorised into four coarse segments, (i) Textual Question Answering (TQA)~\cite{rajpurkar2016squad,rajpurkar2018know,trischler2016newsqa}: question to be answered over a textual paragraph, (ii) Visual Question Answering (VQA)~\cite{antol2015vqa,goyal2017making,johnson2017clevr,krishna2017visual}: question to be answered over an image, (iii) Audio Question Answering (AQA)~\cite{lipping2022clotho,salviclear,fayek2020temporal,behera2023towards}: question to be answered in the context of an audio, and (iv) Video Question Answering (VideoQA)~\cite{lei2018tvqa,deepstory,videoqasurvey}: question to be answered in the context of an video. However, these traditional settings can be achieved by a method that answers question within the input content. As the AI field moving so fast owing to the superior learning and reasoning capabilities of LLMs~\cite{gpt3,falcon,liu2023llava,zhu2023minigpt}, there has been a growing interest in enhancing the question answering capabilities of these assistants to reason over external knowledge beyond the input content. Towards this direction, researchers proposed knowledge-aware benchmarks in textual question answering~\cite{xu2019enhancing,karpukhin2020dense} and visual question answering~\cite{shah2019kvqa,schwenk2022okvqa,infoseek}. Since, AQA in itself being a relatively a new research area, such knowledge-aware question answering benchmark (\data) does not exist in this space. In this paper, we introduce knowledge-intensive audio question answering benchmark and propose three sub-tasks under this task paradigm to fill this research gap encouraging further research in this direction. Further, we propose a framework based on audio entity linking~\cite{li2024ted}, which can be used with any modern day audio-based LMMs to solve knowledge-intensive tasks like \data. 



\section{Audio-based Knowledge Question Answering (\data{}) Benchmark}
\label{sec:dataset}
We introduce three novel benchmark datasets under \textsc{\data}, namely, single audio question answering (s-AQA), multi audio question answering (m-AQA), and retrieval-augmented audio question answering (r-AQA), to study knowledge-intensive audio question answering. These datasets contain 14702, 500, and 115 samples, respectively. More detailed statistics of these datasets are provided in Table~\ref{minitab:s-AQA stats}, Table~\ref{minitab:m-AQA stats}, and Table~\ref{minitab:r-AQA stats} respectively. Data curation for these sub-tasks is discussed next.

\subsection{Data collection}
We leverage the knowledge base (KB) provided by the authors of~\cite{singh2019strings}, specifically we use business-KB provided here\footnote{\href{https://textkvqa.github.io}{https://textkvqa.github.io}} to curate the data for all our three tasks. This KB which is constructed from Wikidata, where each knowledge fact is a triplet connecting two entities with a relation. An example of these triplets is: \textit{Subway, established in, 1965}.  Data collection for three sub-tasks is as follows: (i) \textbf{s-AQA}: Here, we sample triplets with same subject entity and frame an input sentence from these triplets except one triplet which shall be used to frame the question. Question is framed from this excluded triplet of form (subject, relation, object), e.g., for an excluded triplet $<$ \textit{Subway}, \textit{established in}, \textit{1965} $>$, we frame a question \textit{Q: When was Subway established in?}, with the answer being the object in the triplet, \textit{A: 1965} in this case. Input sentence is framed using different triplets with same subject entity, for the above example, we chose the triplet $<$ \textit{Subway}, \textit{serves}, \textit{salad and sandwich} $>$ for creating the input sentence \textit{i/p: Subway serves salad and sandwich}. Input sentence is converted to audio using the text-to-speech model Tachotron 2~\cite{shen2018natural}. (ii) \textbf{m-AQA}: In this task, input consists of two audios, hence, we chose two triplets with different subject entities. For this task, we consider binary answers `Yes' and `No'. For `Yes' answer questions, we sample two triplets  with same (relation, object) but with different subject entities, e.g., consider two triplets $<$ \textit{Subway}, \textit{established in}, \textit{1965} $>$ and $<$ \textit{Arby's}, \textit{established in}, \textit{1964} $>$, we frame a question as follows: \textit{Q: Are these restaurants established in the same decade?}, with answer being \textit{A: Yes}. Similarly,  for `No' answer questions, we sample two triplets with same relation but with different (subject, object) entities, e.g., consider two triplets $<$ \textit{Hotto Motto}, \textit{origin country}, \textit{Japan} $>$ and $<$ \textit{Krispy Kreme}, \textit{origin country}, \textit{United States} $>$, we frame a question as follows: \textit{Q: Are these Japanese restaurants?}, with answer being \textit{A: No}. For both of these cases, input sentences are framed and converted to audio format in a similar fashion as we did in s-AQA. (iii) \textbf{r-AQA}: In this task, input consists of $n$ audios where only few audio samples are relevant to answer the question where as the other audio samples are irrelevant. For positive audio samples, collection is similar to m-AQA. For negative audio samples, we sample entirely different triplets with different (subject, relation, object)s to frame negative sentences, which are converted to audio format in a same way as s-AQA.


\section{Proposed methodology}
\label{sec:method}
\subsection{Audio Entity Linking (AEL)}
\label{sec_sub:AEL}
AEL refers to the task of identifying the mentioned named entities in the given audio and link them to their associated knowledge in an external knowledge base. External knowledge base consists of $n$ named entities $\{N_1, N_2, .., N_n\}$ along with their associated knowledge $\{K_1,K_2, .., K_n\}$. Since, knowledge associated with each entity is a set of triplets, we first frame sentences out of these triplets by simply concatenating subject, relation and object for every entity. Next, we encode the associated knowledge $K_i$, where $i \in \{1,2,..,n\}$ using a text encoder, to obtain knowledge embeddings $E_T^{\{1,2,..n\}}$. Given an audio $A$, the task is to identify and link the mentioned in to one of the named entities in the knowledge base. We first transcribe the audio using wav2vec 2.0~\cite{baevski2020wav2vec2} and encode the transcribed text to obtain the embedding $A_T$. We then compute the cosine similarity of $A_T$ with each of $E_T^{\{1,2,..n\}}$, and choose the knowledge corresponding to the highest similar entity $K_l$ as the linked knowledge. (Illustrated in Fig.~\ref{fig:main_arch} (d)).

\subsection{Knowledge-Augmented Audio Large Multimodal model}
\label{sec_sub:KALM}
We propose a generic framework (\model{}) that can be easily integrated into any LALM. Given an audio input $A$ and a knowledge-intensive question $Q$, our framework enhances the standard LALM prompt. Typically, an LALM would process $Q$ within an instruction prompt alongside audio $A$ to generate an answer. Our approach infuses the instruction prompt with knowledge $K_l$, obtained through AEL of the given audio $A$. This enhanced prompt is then fed into the LALM, enabling it to generate an accurate final answer $a$, leveraging the augmented additional knowledge.


    

\subsection{s-AQA problem formulation and method}
Given a natural language question $Q$ over an input audio $A$, the task is to generate the answer $a$ to the question by reasoning over an external knowledge associated with the named entity mentioned in the audio. Firstly, we perform AEL on the audio $A$ to obtain the linked knowledge $K_l$; and then we feed it along with $Q$ and $A$ to \model{} to obtain the final answer $a$. (Illustrated this in Fig.~\ref{fig:main_arch} (a)).

\subsection{m-AQA problem formulation and method}
Given a natural language question $Q$ over multiple audio samples $A = \{A_1, .., A_k\}$, where $k$ is the number of audio samples, the task is to generate the answer $a$ by reasoning over knowledge associated with named entities mentioned across multiple audios. Firstly, we perform AEL for the input audios and obtain the linked knowledge $K_l^{i}$ for every $i \in \{1, .., k\}$; and concatenate $K_l^{i}$ to obtain $K_l$. Then, we feed $K_l$ along with $Q$, and concatenated audio samples $A$ to \model{} to obtain the final answer $a$. (Illustrated this in Fig.~\ref{fig:main_arch} (b)).


\subsection{r-AQA problem formulation method}
Given a natural language question $Q$ over a set of audio samples $A = \{A_1, .., A_k\}$, where only few audio samples $r$ out of $k$ are relevant, the task is to identify and retrieve the relevant audio samples $A_r$, where $r \in \{1, .., r\}$; once retrieved, the task is similar to m-AQA, to generate answer $a$ by reasoning over the knowledge associated with the mentioned named entities across the retrieved audio samples. Firstly, we perform retrieval, as follows: we obtain ASR-ed text of all the samples and encode them using a text encoder to obtain their embeddings $A_T^{i}$, where $i \in \{1, .., k\}$. Similarly, we obtain the question embedding $Q_T$ using the same text encoder. Next, we compute cosine similarity for $Q_T$ with every $A_T^{i}$, and pick those samples with similarity scores greater than the threshold as the retrieved samples. Once retrieved, we perform AQA using the same methodology as m-AQA to obtain the final answer $a$. (Illustrated in Fig.~\ref{fig:main_arch} (c)).

\begin{table}[t]
  \centering
  \caption{Audio Question Answering Results.}
  \resizebox{\columnwidth}{!}{
  \begin{tabular}{c c c c c c }
\hline
 & & & \multicolumn{3}{c}{Accuracy on \data{}} \\
 \cmidrule(r){4-6}
  Model & Knowledge & AEL+\model{} & s-AQA & m-AQA & r-AQA \\
    \cmidrule(r){1-6}
      \multirow{3}{*}{Audio-flamingo~\cite{audioflamingo}} & \xmark & \xmark  & 0.09 & 50.4 & 17.2 \\
  & \cmark  & \cmark  & \textbf{15.3} & \textbf{57.6} & \textbf{21.5} \\
  \cdashline{2-6}
  & \cmark  & GT & 22.8 & 60.8 & 23.3 \\
   \cmidrule(r){2-6}
  \multirow{3}{*}{GAMA~\cite{ghosh2024gama}} & \xmark & \xmark & 0.06 & 14.8 & 5.1 \\
     & \cmark & \cmark & \textbf{31.4} & \textbf{35.2} & \textbf{13.1} \\
       \cdashline{2-6}
  & \cmark  & GT  & 51.1 & 38.8 & 14.9 \\
     \cmidrule(r){2-6}
  \multirow{3}{*}{LTU-AS~\cite{ltu}} & \xmark  & \xmark & 3.9 & 48.6 & 16.5 \\
     & \cmark & \cmark & \textbf{54.7} & \textbf{55.2} & \textbf{20.3} \\
       \cdashline{2-6}
  & \cmark  & GT & 76.2 & 56 & 21.5 \\
    \hline
    
\end{tabular}}
\label{tab:mainRes-1}
\end{table}

\begin{table}[t]
  \centering
  \scriptsize
  \caption{Audio Entity Linking Results.}
  {
  \begin{tabular}{c c c c c}
  \hline
   Knowledge source & s-AQA (Acc.) & m-AQA (Acc.) & r-AQA (F1) \\
 \cmidrule(r){1-5}
  Entity-name & 56.1 & 55.6 & 66.7 \\
  20\% knowledge & 58 & 55.2 & 68 \\
  Full knowledge & \textbf{63.9} & \textbf{59.1} & \textbf{68} \\
  \hline
    
\end{tabular}}
\label{tab:ael_results}
\end{table}

\section{Experiments, Results and Analysis}

\noindent\textbf{Metrics}: We use accuracy to evaluate the models. For AQA, accuracy is 1 if the ground truth answer is present in the generated text, and 0 otherwise. For AEL in s-AQA and m-AQA, accuracy is 1 if entities are correctly predicted. For r-AQA, which involves retrieval, we use the F1 score. \textbf{Results and Discussion}: We assess our proposed framework (AEL + \model{}) across three recent LALMs: GAMA~\cite{ghosh2024gama}, LTU-AS~\cite{ltu}, and Audio-flamingo~\cite{audioflamingo}, on three sub-tasks of \data{}, reporting AQA accuracy in Table~\ref{tab:mainRes-1}. s-AQA, which requires open-ended answers, is the most challenging, highlighting the importance of external knowledge provided by \model{}. m-AQA and r-AQA with finite answers, are simpler. Despite extensive multimodal pretraining, these LALMs struggle with knowledge-intensive tasks, particularly open-ended answer generation. The table results demonstrate that knowledge augmentation via \model{} significantly improves performance across all sub-tasks. Table~\ref{tab:ael_results} shows AEL results under different conditions: (i) entity-alone, where only the entity name is used for AEL, (ii) partial knowledge, where only 20\% of the knowledge is used, and (iii) full knowledge. This indicates that any form of knowledge improves performance, with full knowledge being the most effective. We also present an upper bound on AQA accuracy in Table~\ref{tab:mainRes-1} by showing results in an oracle setting, assuming 100\% accurate entity linking and using knowledge from the ground truth (GT) entity. Even in this ideal scenario, LALMs fall short, indicating that the problem remains unsolved. All experiments are conducted in a zero-shot setting with no fine-tuning. From our findings, LTU-AS~\cite{ltu} is the best-performing LALM (54.7 on s-AQA, 55.2 on m-AQA and 21.5 on r-AQA). To investigate the impact of full knowledge in AEL on AQA performance, we substitute full knowledge with (i) entity name and (ii) partial knowledge in AEL and evaluate AQA using LTU-AS. We report these results in Table~\ref{tab:ael_results} show that full knowledge text is crucial for better AQA performance, as accuracy drops to 49.3 from 54.7 without it. \textbf{Limitations}: We note the following limitations in our work: (i) The dataset curation process assumes that each audio contains only one named entity mention, which may not reflect real-world scenarios. (ii) It is assumed that all mentioned entities are in English, which may not hold true in practical settings. (iii) As a curated synthetic dataset, we have manually quantified that a small fraction of samples (less than 5\%) contain noise.

\begin{table}[!t]
  \centering
  \caption{Results on s-AQA with various knowledge using LTU-AS~\cite{ltu}.}
  {
    \begin{tabular}{l c }
    \hline
    Knowledge source & \multicolumn{1}{c}{Acc.} \\ 
    \hline
    - & 0.06 \\ 
    Entity-name & 49.3 \\
    20\% knowledge & 51.3  \\
    Full knowledge & \textbf{54.7} \\
    \hdashline
    Oracle & 76.2 \\
    \hline
    \end{tabular}
  }
  \label{tab:know_source_ablation}
\end{table}

\section{Conclusion and Future Work}
We have drawn the attention of the audio community towards a novel task \textit{\data{}: Audio Question Answering with Knowledge} by introducing three knowledge-intensive sub-tasks, namely, s-AQA, m-AQA and r-AQA respectively. We have also proposed a generic framework (AEL+\model{}) that can be easily adapted to any modern day LALM to equip them with knowledge reasoning capabilities required to perform tasks like \data{}. 
We strongly believe that an ideal model for \data{} should naturally excel in both audio understanding and knowledge reasoning without needing separate entity linking and knowledge augmentation. This dataset helps evaluate and benchmark existing and future models on these essential capabilities. The future scope of this work is to expand the dataset by enhancing entity coverage across diverse domains and addressing similar tasks in multilingual contexts.

\bibliographystyle{IEEEbib}
\bibliography{refs}
\end{document}